\pdfoutput=1

\documentclass[11pt]{article}

\usepackage[]{ACL2023}

\usepackage{times}
\usepackage{latexsym}

\usepackage[T1]{fontenc}

\usepackage[utf8]{inputenc}
\usepackage{color}
\usepackage{tabularx}
\usepackage{tabu}
\usepackage{booktabs}
\usepackage{multirow}

\usepackage{graphicx} 
\usepackage{float}
\usepackage{subfigure} 
\usepackage{amssymb}

\usepackage{hyperref}
\usepackage{tablefootnote}
\usepackage{xspace}

\usepackage{float}
\usepackage{amsmath}
\usepackage{bm}
\usepackage{algorithmicx}
\usepackage{algorithm}
\usepackage{algpseudocode}
\usepackage{makecell}

\usepackage{microtype}

\usepackage{enumitem}
\usepackage{arydshln}
\usepackage{colortbl} 
\usepackage{xcolor}

\newcommand{\ours}{\textsc{RobuT}\xspace}
\newcommand{\framework}{\textsc{LeTA}\xspace}
\newcommand{\tableqa}{Table QA\xspace}

\newcommand{\tapas}{\textsc{TaPas}\xspace}
\newcommand{\wikisql}{\textsc{WikiSQL-Weak}\xspace}
\newcommand{\wikisqlshort}{\textsc{WikiSQL}\xspace}
\newcommand{\num}{10\xspace}
\newcommand{\totalnum}{143,477\xspace}

\newcommand{\red}[1]{\textcolor{red}{#1}}
\newcommand{\blue}[1]{\textcolor{blue}{#1}}

\title{\ours: A Systematic Study of Table QA Robustness Against \\Human-Annotated Adversarial Perturbations} 
\author{Yilun Zhao$^1$ \quad Chen Zhao$^2$ \quad Linyong Nan$^1$ \quad Zhenting Qi$^3$ \quad Wenlin Zhang$^3$ \\
\bf{\quad Boyu Mi$^3$ \quad Xiangru Tang$^1$ \quad Dragomir Radev$^1$} \\
$^1$Yale University \quad $^2$ New York University \quad $^3$ Zhejiang University \\
\texttt{yilun.zhao@yale.edu \quad cz1285@nyu.edu}
}

\begin{document}
\maketitle
\begin{abstract}
Despite significant progress having been made in question answering on tabular data (Table QA), it's unclear whether, and to what extent existing Table QA models are robust to task-specific perturbations, e.g., replacing key question entities or shuffling table columns.
To systematically study the robustness of Table QA models, we propose a benchmark called \ours, which builds upon existing Table QA datasets (WTQ, \wikisql, and SQA) and includes human-annotated adversarial perturbations in terms of table header, table content, and question.
Our results indicate that both  state-of-the-art Table QA models and large language models (e.g., GPT-3) with few-shot learning falter in these adversarial sets.
We propose to address this problem by using large language models to generate adversarial examples to enhance training, which significantly improves the robustness of Table QA models. 
Our data and code is publicly available at \url{https://github.com/yilunzhao/RobuT}.

\end{abstract}

\section{Introduction}

Table QA uses structured table as world knowledge to answer questions.
In recent years, Transformer-based models~\cite{yin-etal-2020-tabert, herzig-etal-2020-tapas, yang-etal-2022-tableformer, jiang-etal-2022-omnitab, liu2022tapex,scao2022bloom} achieve remarkable results on existing \tableqa benchmark datasets~\cite{pasupat-liang-2015-compositional, zhongSeq2SQL2017, iyyer-etal-2017-search}.
Despite significant progress, state-of-the-art models are only evaluated within the same distribution, which does not provide insight into the model's robustness against out-of-domain distribution or adversarial data~\cite{suhr-etal-2020-exploring}, and recent studies~\cite{Cho2018AdversarialTA, 2005.12696, yang-etal-2022-tableformer} revealed that existing models are vulnerable to adversarial perturbations. 
For example, \citet{Cho2018AdversarialTA} observed significant performance degradation after a sentence-level question perturbation. \citet{yang-etal-2022-tableformer} showed that state-of-the-art \tableqa models exhibited a dramatic performance drop after randomly shuffling the row or column order of the input table.  
However, previous works primarily focus on a single type of adversarial perturbation and rely on rule-based perturbation methods that are limited in linguistic richness. 
We fill this gap through a comprehensive evaluation of \tableqa model robustness. 

\begin{figure}[!t]
    \centering
    \includegraphics[width = \linewidth]{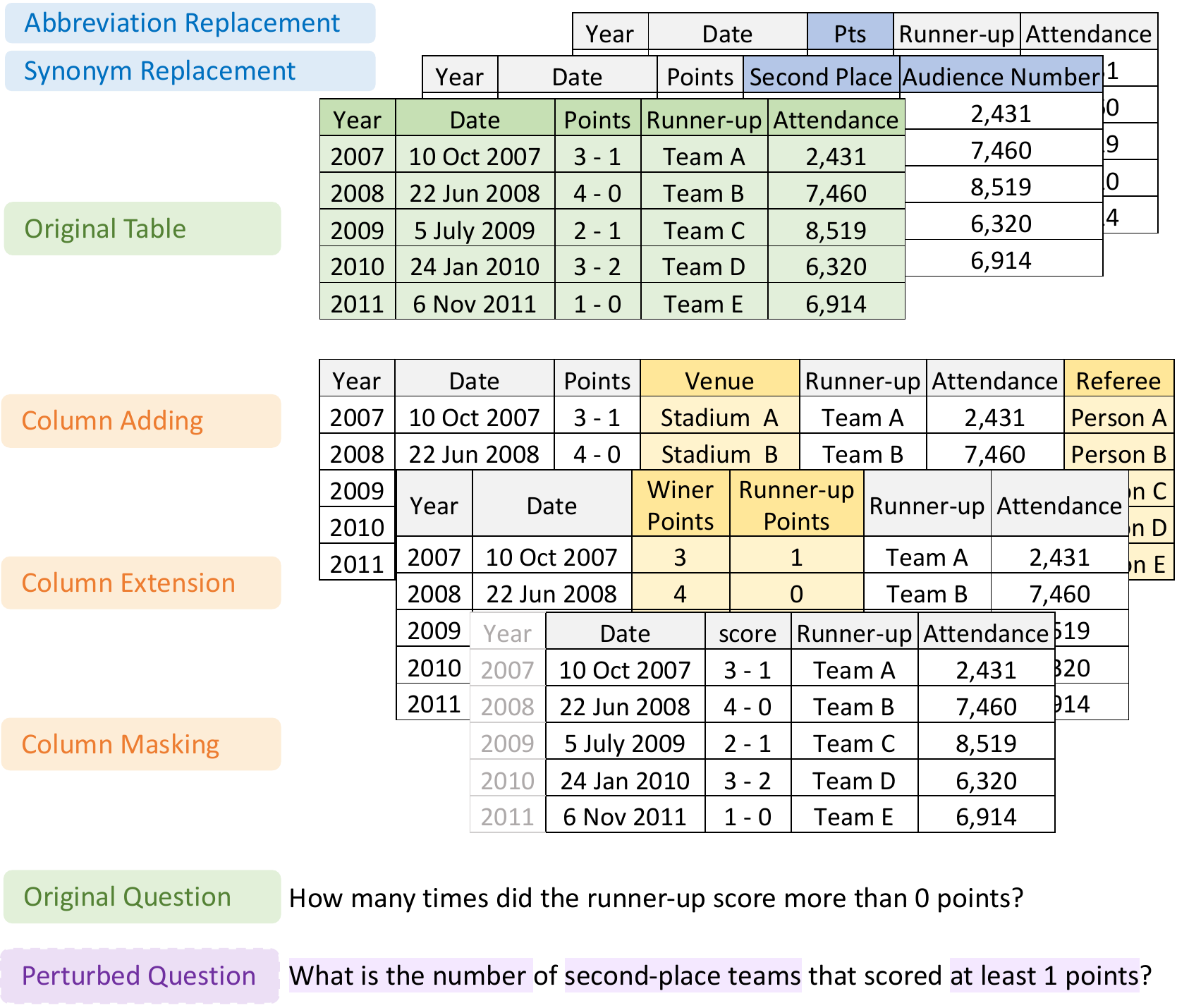}
    \caption{Examples of adversarial perturbation over table header (blue), table content (orange), and question (purple). Table QA model predicts a correct answer on the original example but fails on perturbed ones. }
    \label{fig:main_example}
\end{figure}
In this paper, we constructed a new benchmark, \textbf{\ours}, to systematically evaluate the \textbf{\textsc{Robu}}stness of \textbf{T}able QA models (Figure~\ref{fig:main_example}). \ours was built upon the development set of WTQ~\cite{pasupat-liang-2015-compositional}, \wikisql~\cite{zhongSeq2SQL2017}, and SQA~\cite{iyyer-etal-2017-search} datasets. 
Specifically, we designed \num types of adversarial perturbations at three different levels (i.e., table header, table content, and natural language question), with a total number of \totalnum \emph{human-annotated} perturbed examples.

We evaluated state-of-the-art \tableqa models~\cite{herzig-etal-2020-tapas,  chen2021evaluating, liu2022tapex, yang-etal-2022-tableformer, jiang-etal-2022-omnitab, table-cot} and few-shot learning with large language models (LLMs) on \ours. The experiments revealed that all models significantly degrade performance in our adversarial sets, while large LLMs, such as GPT-3~\cite{gpt-3, cot} and CodeX~\cite{chen2021evaluating}, are more robust. 
For example, GPT-3 outperforms all other Table QA models on both word-level and sentence-level question perturbations.

Motivated by the findings that LLMs are more robust against human-annotated adversarial perturbations, we developed \textbf{\framework}, a \textbf{L}LM-\textbf{\textsc{e}}nhanced \textbf{T}able QA \textbf{A}ugmentation framework that uses LLMs to generate adversarial examples to enhance model training. Specifically, we prompted GPT-3 or CodeX to simulate human annotation and generate adversarial training examples for all perturbation types. Experimental results showed that fine-tuning on these adversarial training examples significantly improves model robustness.

We summarize three major contributions:
\begin{itemize}
    \item We constructed \ours, the first diagnostic evaluation benchmark for \tableqa robustness. We applied rigid annotation quality control procedure to ensure the comprehensiveness, linguistic richness, and semantic association of the benchmark.
    \item Experimental results showed that state-of-the-art models exhibited significant performance drops on \ours benchmark, thus there is still large room to explore for \tableqa tasks beyond high leaderboard scores.
    \item We designed \framework, an adversarial training example generation framework using LLM prompting methods. Experiments demonstrated that our methods effectively improves \tableqa model robustness. 
\end{itemize}

\begin{table*}[!t]
\centering
\small
\begin{tabular}{llrr}
    \toprule
    Dataset & Type & \# Tables & \# Examples \\
    \midrule
    WTQ~\cite{pasupat-liang-2015-compositional} & Complex QA & 2,108 & 22,033\\
    \wikisql~\cite{zhongSeq2SQL2017} & Simple QA & 24,241 & 80,654\\
    SQA~\cite{iyyer-etal-2017-search} & Conversational QA & 982 & 6,066\\

    \bottomrule
\end{tabular}

\caption{An overview of the WTQ, \wikisql, and SQA datasets.}
\label{tab:tableqa_stat}
\end{table*}
\section{Related Work}
\paragraph{Table QA}
Question answering over tables has received significant attention as it helps non-expert users interact with complex tabular data.
This problem is originally framed as semantic parsing, also known as Text-to-SQL parsing~\cite{yu-etal-2018-spider, yu-etal-2019-sparc, wang-etal-2020-dusql, guo-etal-2021-chase}, in which the parser takes both question and table header as input, and predicts a SQL query that is directly executable to get the answer. However, 
training state-of-the-art Text-to-SQL parsers require large amounts of expensive SQL annotations, limiting its applicability to real scenarios; In addition, these Text-to-SQL parsers make a simplified assumption that only table headers are necessary while ignoring the value of table contents. 
To mitigate these issues, recent works ignore generating SQL queries, and instead follow \emph{retrieve then reason} paradigm~\cite{yin-etal-2020-tabert, herzig-etal-2020-tapas, eisenschlos-etal-2020-understanding, yang-etal-2022-tableformer, liu2022tapex, jiang-etal-2022-omnitab, zhao-etal-2022-reastap}, which first retrieve information from the table, and conduct human-like reasoning to answer the question. 
With the help of pre-training on large scale table corpus, these approaches have achieved remarkable results on several \tableqa benchmarks, including WikiTableQuestions~\cite{pasupat-liang-2015-compositional}, \wikisql~\cite{zhongSeq2SQL2017}, and SQA~\cite{iyyer-etal-2017-search}. 
More recently, \citet{table-cot} found that LLMs~\cite{gpt-3, chen2021evaluating} with few-shot in-context learning shows promise on the Table QA task.

\paragraph{Robustness in Table-Relevant Task}
Assessing model robustness is crucial for building trustworthy models~\cite{wang2021adversarial, chang-etal-2021-robustness, goel-etal-2021-robustness, wang-etal-2022-identifying, wang-etal-2022-measure, gupta-etal-2022-right}. Recent work~\cite{gan-etal-2021-towards, zeng-etal-2020-photon, anonymous2023drspider} has focused on evaluating the robustness of text-to-SQL parsing models, and designed test sets with perturbations including NLQ input, table headers, and SQL queries. A major limitation is that these perturbations (e.g., lexical substitutions) are often targeted at a vulnerable key component that is specific to text-to-SQL parsing: schema linking~\cite{wang-etal-2020-rat, scholak-etal-2021-picard}, which matches table headers question keywords.
Our study is focused on Table QA in general, and we make two key differences: First, in addition to existing perturbations, we also perturbed \textit{table contents}, valuable information that is often dismissed by Text-to-SQL models. Second, unlike previous works that used human to verify perturbations generated from heuristics or models, we directly adopted human-annotated perturbations to ensure high data quality.

\paragraph{Adversarial Data Generation}
Existing works have proposed data augmentation and adversarial training techniques to improve model robustness. In the field of table-relevant tasks, \citet{gan-etal-2021-towards} applied the BERT-Attack model~\cite{li-etal-2020-bert-attack} to generate adversarial training questions to improve the Table QA model's robustness against synonym substitution. \citet{pi-etal-2022-towards} and \citet{zhao-etal-2022-bridging} proposed to train Table QA models over examples with perturbed database schema to defend schema-level adversarial attack.  
Recent approaches applied LLMs~\cite{gpt-3, opt} to generate adversarial data.
For example, the evaluation data for NLQ-level perturbation in the Dr.Spider benchmark~\cite{anonymous2023drspider} were generated using LLM-prompting methods~\cite{DBLP:journals/corr/abs-2107-13586, bach2022promptsource}. In contrast, we created our test sets through human annotation, and applied LLMs to generate adversarial \textit{training} examples to enhance training Table QA models.

\section{\ours Benchmark}

We constructed \ours to comprehensively evaluate the robustness of Table QA models against task-specific adversarial perturbations annotated by human experts.
To ensure the high annotation quality of \ours benchmark, 
we first designed the following three \emph{annotation principles}: 
\begin{itemize}[parsep=0pt,leftmargin=*]
    \item \textbf{Diagnostic Comprehensiveness:} To provide a comprehensive study,  the benchmark should enumerate different diagnostic angles over multiple task-specific perturbation categories.
    \item \textbf{Phraseology Correctness and Richness:} The perturbations should follow linguistic phraseology conventions and are linguistically rich, which cannot be achieved by rule-based or model-based methods.
    \item \textbf{Semantic Association:} The perturbed part should still maintain the meanings of the original contexts, e.g., the new table should maintain the same domain after adding a few columns.
\end{itemize}

Following the aforementioned annotation principles, we curated \ours based on the \emph{development set}\footnote{For WTQ and SQA datasets that have multiple official train/dev splits for the purpose of cross-validation, we used the split of \texttt{random-split-1-\{train/dev\}} in our work.} of three mainstream Table QA datasets: WTQ~\cite{pasupat-liang-2015-compositional}, which contains human-annotated questions over Wikipedia tables and requires complex reasoning; Weakly-supervised \textsc{WikiSQL}~\cite{zhongSeq2SQL2017}, which requires models to filter and optionally aggregate on table cell values to obtain the answer; and SQA~\cite{iyyer-etal-2017-search}, in which annotators decompose questions originally from WTQ to a sequence of questions (2.9 questions per sequence on average).
The statistics of these three Table QA datasets are shown in Table~\ref{tab:tableqa_stat}.

We designed a total of \num perturbation types on four different levels (i.e., table header, table content, natural language question, and mix). 
And as we have three subsets, our final dataset includes 30 \emph{test sets} in total.
Each test set contains parallel pre-perturbation
and post-perturbation data to measure model robustness against the perturbation. In total, \ours
contains \totalnum pairs of examples, including 39,471 examples from \ours-\textsc{WTQ}, 83,816 examples from \ours-\textsc{WikiSQL}, and 14,862 examples from \ours-SQA. 

\subsection{Table Header Perturbation}

Table QA models often match the question segments to the table header in order to identify the relevant columns. However, most examples in existing Table QA datasets only consist of \emph{exact match} scenarios~\cite{suhr-etal-2020-exploring}, leaving it unclear if models can handle table header variations.
The goal of table header perturbation is to replace some column names of the table header with their \emph{synonyms} or \emph{abbreviations} that might mislead existing Table QA models.

\paragraph{Header Synonym Replacement} Given a table, the annotators were asked to first identify the columns that can be renamed. For each candidate column, they were required to come up with a synonymous column name that maintains the same domain-relevancy. 
For example, the column ``\texttt{runner-up}" in a table about sports can be renamed as ``\texttt{second place}". The annotators were given full access to a public synonym website\footnote{\url{https://www.thesaurus.com/}} as the reference of the synonymous names. 

\paragraph{Header Abbreviation Replacement} For each table, we first collected abbreviation(s) of its column names, using APIs provided by a public abbreviation website\footnote{\url{https://www.abbreviations.com/}}. 
The abbreviation would replace the original column name if the annotators decided that it is appropriate for the given table context.

\subsection{Table Content Perturbation}\label{sec:table_content_perturb}

To answer the given question, 
Table QA models should understand table contents, retrieve relevant cells, and reason over them.
However, \citet{yang-etal-2022-tableformer} has found that existing Table QA models learn unwanted bias related to the table contents. 
In our preliminary work, we also found that questions in WTQ often use information from the first three or last two rows of the table as the answer. This finding suggests that the existing Table QA datasets actually contain annotation bias related to table content, as annotators are more likely to compose questions for the first or last few rows of the table. 
To evaluate the Table QA model robustness against table content variation, we designed five perturbation types to alter the table content in column-level or row-level that do not affect the final answers.

\paragraph{Row Order or Column Order Shuffling} For each table, we randomly shuffled the order of its rows or columns. We excluded a small number of questions asking about the absolute table position since their answers will change after shuffling (e.g., ``what is the last region listed on the table?"). 

\paragraph{Column Extension} Column extension perturbation extends existing columns, including column name and column content, into multiple semantic-equivalent columns. 
Instead of using rule-based methods~\cite{zhao-etal-2022-bridging}, we asked annotators to provide possible semantically equivalent substitutions for each column. Specifically, they were asked to decompose a compound column into multiple columns, such as replacing the column ``\texttt{Score}” in a table about soccer games with ``\texttt{Home Team Score}" and ``\texttt{Away Team Score}".

\paragraph{Column Masking}
Some table columns are correlated to each other. For example, the column ``\texttt{Ranking}” can be inferred by another column ``\texttt{Total Points}”. We asked the annotators to mask the columns whose content could be inferred by other columns.

\paragraph{Column Adding}\label{sec:dense_retrieve}
Column adding perturbs table content by introducing new columns that are semantically associated with the original table context. Following \citet{pi-etal-2022-towards}, for each table, we applied the TAPAS-based dense retriever~\cite{herzig-etal-2020-tapas} to retrieve the most relevant tables from Web Data Commons database~\cite{webdatabase}. We collected the three most relevant tables for each source table. The annotators were then asked to follow the \emph{semantic-association} annotation principle, and select some columns that can be randomly inserted into the original table.

\begin{table*}[!t]
\setlength\tabcolsep{3pt}
\centering
\resizebox{0.98\textwidth}{!}{%
\begin{tabular}{llrc*{4}{ccc}cc}
\toprule
\multirow{2}{*}{\begin{tabular}[c]{@{}l@{}}Level\end{tabular}} & \multirow{2}{*}{\begin{tabular}[c]{@{}l@{}}Perturbation Type\end{tabular}} & \multirow{2}{*}{\begin{tabular}[c]{@{}l@{}}\# Example\end{tabular}} && \multicolumn{2}{c}{TAPAS} && \multicolumn{2}{c}{TableFormer} && \multicolumn{2}{c}{TAPEX} && \multicolumn{2}{c}{OmniTab} && \multicolumn{2}{c}{GPT-3} \\ 
\noalign{\vskip 0.5ex}\cline{5-6} \cline{8-9} \cline{11-12} \cline{14-15} \cline{17-18}\noalign{\vskip 0.5ex}

 & & & &\textsc{Acc}& \textsc{R-Acc} && \textsc{Acc}& \textsc{R-Acc} &&\textsc{Acc}& \textsc{R-Acc}&&\textsc{Acc}& \textsc{R-Acc} &&\textsc{Acc}& \textsc{R-Acc} \\
\midrule
 &  Development Set & 2,831 && 48.3 & -- && 51.3 & -- && 57.3 & -- && 61.0 & -- && 42.9 & -- \\
\midrule
 
\multirow{4}{*}{\begin{tabular}[c]{@{}l@{}}Table\\Header\end{tabular}} 
& \multirow{2}{*}{\begin{tabular}[c]{@{}l@{}}Synonym Replacement\end{tabular}} &
\multirow{2}{*}{4,185} && 44.7 / 38.5 & \multirow{2}{*}{81.1} & & 47.0 / 41.1 & \multirow{2}{*}{83.2} & & 54.3 / 48.4 & \multirow{2}{*}{84.6} & & 58.5 / 54.0 & \multirow{2}{*}{\underline{88.0}} & & 41.7  /  39.9 & \multirow{2}{*}{\textbf{90.7}} \\
& & & & \color{red}{(-6.2)} && & \color{red}{(-5.9)} && & \color{red}{(-5.9)} && & \color{red}{(-4.5)} && & \color{red}{(-1.8)} &\\
\noalign{\vskip 0.5ex}\cdashline{2-18}\noalign{\vskip 0.5ex}

& \multirow{2}{*}{\begin{tabular}[c]{@{}l@{}}Abbreviation Replacement\end{tabular}} &
\multirow{2}{*}{2,878} && 43.4 / 35.1 & \multirow{2}{*}{76.1} & & 45.3 / 37.1 & \multirow{2}{*}{76.9} & & 50.4 / 44.3 & \multirow{2}{*}{83.7} & & 54.8 / 52.0  & \multirow{2}{*}{\underline{89.5}} & & 41.5  /  39.2 & \multirow{2}{*}{\textbf{93.8}} \\
& & & & \color{red}{(-8.3)} && & \color{red}{(-8.2)} && & \color{red}{(-6.1)} && & \color{red}{(-2.8)} && & \color{red}{(-2.3)} &\\
\midrule

\multirow{10}{*}{\begin{tabular}[c]{@{}l@{}}Table\\Content\end{tabular}} 
& \multirow{2}{*}{\begin{tabular}[c]{@{}l@{}}Row Order Shuffling\end{tabular}} &
\multirow{2}{*}{7,636} && 48.0 / 40.6 & \multirow{2}{*}{74.8} & & 51.0  /  50.9 & \multirow{2}{*}{\textbf{97.0}} & & 56.9 / 45.7 & \multirow{2}{*}{71.7} & & 60.6 / 51.2 & \multirow{2}{*}{77.8} & & 42.9  /  38.5 & \multirow{2}{*}{\underline{90.2}} \\
& & & & \color{red}{(-7.4)} && & \color{red}{(-0.1)} && & \color{red}{(-11.2)} && & \color{red}{(-9.4)} && & \color{red}{(-4.4)} &\\
\noalign{\vskip 0.5ex}\cdashline{2-18}\noalign{\vskip 0.5ex}

& \multirow{2}{*}{\begin{tabular}[c]{@{}l@{}}Column Order Shuffling\end{tabular}} &
\multirow{2}{*}{6,508} && 45.7 / 42.5 & \multirow{2}{*}{86.5} & & 51.2  /  51.0 & \multirow{2}{*}{\textbf{99.1}} & & 54.4 / 48.5 & \multirow{2}{*}{81.4} & & 58.4 / 56.0 & \multirow{2}{*}{89.2} & & 40.9  /  40.0 & \multirow{2}{*}{\underline{93.3}} \\
& & & & \color{red}{(-3.2)} && & \color{red}{(-0.2)} && & \color{red}{(-5.9)} && & \color{red}{(-2.4)} && & \color{red}{(-0.9)} &\\
\noalign{\vskip 0.5ex}\cdashline{2-18}\noalign{\vskip 0.5ex}

& \multirow{2}{*}{\begin{tabular}[c]{@{}l@{}}Column Extension\end{tabular}} &
\multirow{2}{*}{2,672} && 50.9 / 42.5 & \multirow{2}{*}{73.4} & & 52.5  /  45.0 & \multirow{2}{*}{\underline{74.8}} & & 61.2 / 47.8 & \multirow{2}{*}{71.4} & & 64.5 / 52.9 & \multirow{2}{*}{74.7} & & 43.1  /  37.4 & \multirow{2}{*}{\textbf{81.4}} \\
& & & & \color{red}{(-8.4)} && & \color{red}{(-7.5)} && & \color{red}{(-13.4)} && & \color{red}{(-11.6)} && & \color{red}{(-5.7)} &\\
\noalign{\vskip 0.5ex}\cdashline{2-18}\noalign{\vskip 0.5ex}

& \multirow{2}{*}{\begin{tabular}[c]{@{}l@{}}Column Masking\end{tabular}} &
\multirow{2}{*}{425} && 47.9  /  45.2 & \multirow{2}{*}{91.0} & & 51.0  /  47.7 & \multirow{2}{*}{87.2} & & 56.7  /  54.4 & \multirow{2}{*}{94.6} & & 60.4  /  58.0 & \multirow{2}{*}{\underline{94.9}} & & 42.4 / 41.9 & \multirow{2}{*}{\textbf{97.0}} \\
& & & & \color{red}{(-2.7)} && & \color{red}{(-3.3)} && & \color{red}{(-2.3)} && & \color{red}{(-2.4)} && & \color{red}{(-0.5)} &\\
\noalign{\vskip 0.5ex}\cdashline{2-18}\noalign{\vskip 0.5ex}

& \multirow{2}{*}{\begin{tabular}[c]{@{}l@{}}Column Adding\end{tabular}} &
\multirow{2}{*}{4,574} && 48.9 / 47.1 & \multirow{2}{*}{\textbf{89.3}} & & 51.9 / 48.7 & \multirow{2}{*}{83.5} & & 57.4 / 50.4 & \multirow{2}{*}{80.1} & & 61.6 / 57.2 & \multirow{2}{*}{84.8} & & 41.3  /  36.8 & \multirow{2}{*}{\underline{85.6}} \\
& & & & \color{red}{(-1.8)} && & \color{red}{(-3.2)} && & \color{red}{(-7.0)} && & \color{red}{(-4.4)} && & \color{red}{(-4.5)} &\\

\midrule

\multirow{4}{*}{\begin{tabular}[c]{@{}l@{}}NLQ\end{tabular}} 
& \multirow{2}{*}{\begin{tabular}[c]{@{}l@{}}Word-Level Paraphrase\end{tabular}} &
\multirow{2}{*}{2,346} && 45.6 / 38.6 & \multirow{2}{*}{77.8} & & 49.5  /  42.7 & \multirow{2}{*}{78.5} & & 54.7 / 49.2 & \multirow{2}{*}{84.3} & & 58.0 / 54.1 & \multirow{2}{*}{\underline{86.8}} & & 41.2  /  40.3 & \multirow{2}{*}{\textbf{93.7}} \\
& & & & \color{red}{(-7.0)} && & \color{red}{(-6.8)} && & \color{red}{(-5.5)} && & \color{red}{(-3.9)} && & \color{red}{(-0.9)} &\\
\noalign{\vskip 0.5ex}\cdashline{2-18}\noalign{\vskip 0.5ex}

& \multirow{2}{*}{\begin{tabular}[c]{@{}l@{}}Sentence-Level Paraphrase\end{tabular}} &
\multirow{2}{*}{2,404} && 45.6 / 41.1 & \multirow{2}{*}{80.8} & & 49.6 / 44.0 & \multirow{2}{*}{77.1} & & 54.8 / 49.5 & \multirow{2}{*}{84.0} & & 58.2 / 55.4  & \multirow{2}{*}{\underline{87.0}} & & 41.0  /  40.5 & \multirow{2}{*}{\textbf{94.2}} \\
& & & & \color{red}{(-4.5)} && & \color{red}{(-5.6)} && & \color{red}{(-5.3)} && & \color{red}{(-2.8)} && & \color{red}{(-0.5)} &\\
\midrule

\multirow{2}{*}{Mix} 
& \multirow{2}{*}{--} &
\multirow{2}{*}{3,012} && 44.5  /  32.0 & \multirow{2}{*}{64.7} & & 47.6  /  35.3 & \multirow{2}{*}{63.4} & & 52.0  /  39.5 & \multirow{2}{*}{70.5} & & 64.5  /  43.2 & \multirow{2}{*}{\underline{74.0}} & & 37.4  /  30.6 & \multirow{2}{*}{\textbf{83.2}} \\
& & & & \color{red}{(-12.5)} && & \color{red}{(-12.3)} && & \color{red}{(-12.5)} && & \color{red}{(-11.3)} && & \color{red}{(-6.8)} &\\
\bottomrule
\end{tabular}
}
\caption{Data statistics and robustness evaluation results of state-of-the-art Table QA models on \ours-WTQ. \textsc{Acc} represents the \emph{Pre-} and \emph{Post-perturbation Accuracy}; \textsc{R-Acc} represents the \emph{Robustness Accuracy}. Bold numbers indicate the highest \emph{Robustness Accuracy} in each perturbation type, and underscores denote the second best result. When evaluating GPT-3 (i.e., \texttt{text-davinci-003}) in a few-shot setting, we reported results on 200 randomly sampled examples for each perturbation type. The results of \ours-\wikisqlshort and \ours-SQA are shown in Table~\ref{tab:wikisql_results} and \ref{tab:sqa_results} in Appendix. 
}
\label{tab:wtq_attack_results}
\end{table*}

\subsection{NLQ Perturbation}\label{sec:NLQ_perturb}
\begin{figure}[!t]
    \centering
    \includegraphics[width = \linewidth]{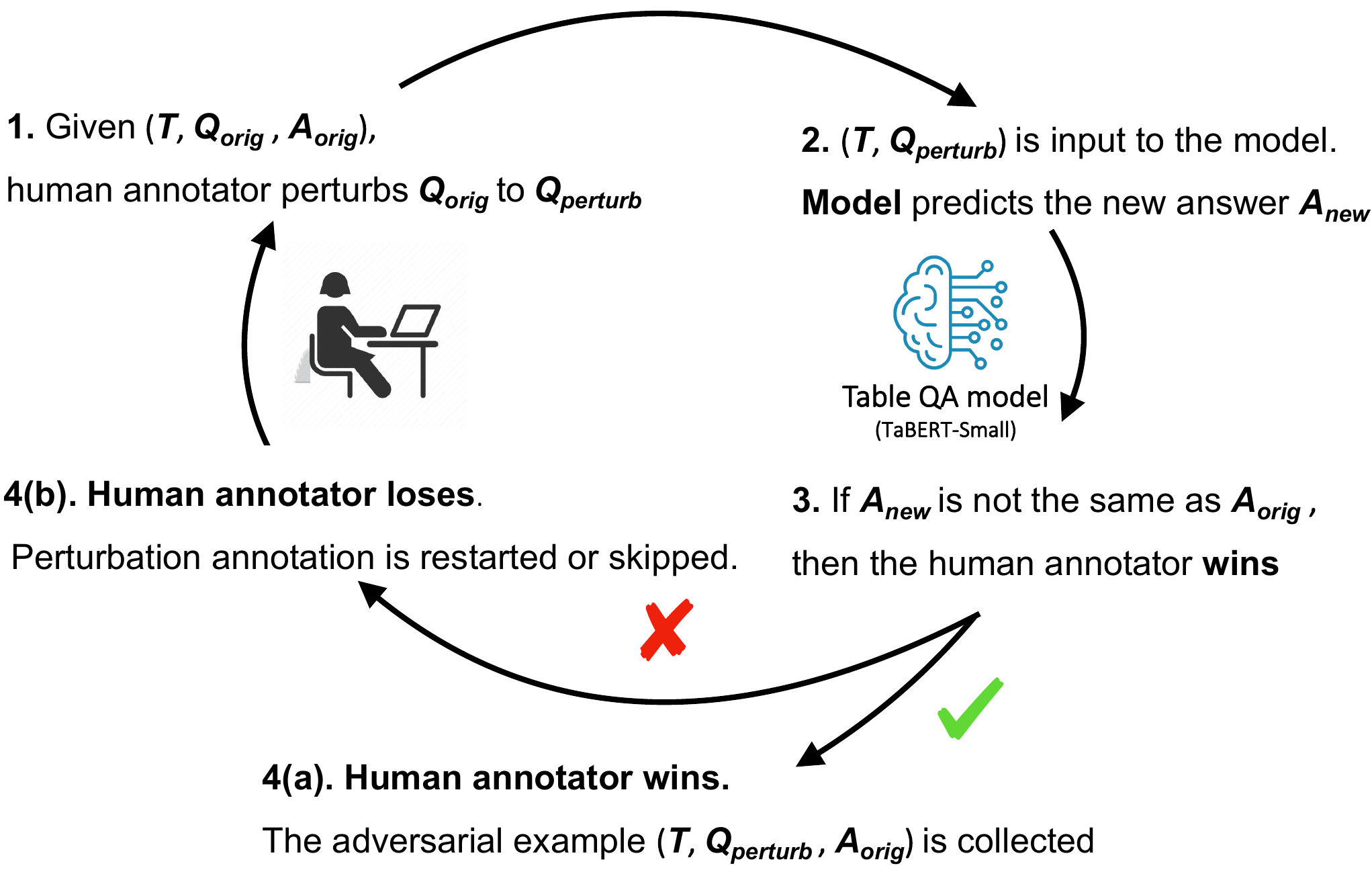}
    \caption{Overview of adversarial annotation process to collect perturbed NLQs in word-level and sentence-level using a model in the loop. $\bm{A_{orig}}$ is the answer predicted by the Table QA model (i.e., TaBERT-small), given the table $\bm{T}$ and pre-perturbed question $\bm{Q_{orig}}$.}
    \label{fig:nlq_annotation}
\end{figure}

In addition to table headers and contents, the input questions also affect model robustness. Our initial analysis 
found that questions from existing datasets  contain annotation bias, causing models to learn shortcuts. 
For example, in WTQ, questions related to \emph{counting} operation usually start with the phrase ``how many”. And if we change the phase to ``what is the quantity of", the fine-tuned models are likely to predict wrong, as they rely on the alignments between ``how many” and \emph{counting} operation.

To systematically evaluate Table QA model robustness against NLQ perturbation, we applied a model-in-the-loop adversarial example annotation framework~\cite{bartolo-etal-2020-beat} to collect new questions perturbed in word-level and sentence-level. 
As shown in Figure \ref{fig:nlq_annotation}, a finetuned TaBERT-small~\cite{yin-etal-2020-tabert} model was integrated into the annotation process. The annotators could directly interact with the model predictions during the annotation process. They were required to perturb questions at the word-level or sentence-level that could change the model's predictions.

\paragraph{Word-level Perturbation}
For word-level NLQ perturbation, we required annotators to focus on perturbing the key entities in the question, such as replacing the entity with its synonym.  

\paragraph{Sentence-level Perturbation}
For sentence-level NLQ perturbation, we required annotators to focus on perturbing the sentence structure, while maintaining its overall meaning. We did not consider the adversarial type of adding noise to the original question as it would change question's meaning. 

\subsection{Mix Perturbation}
In previous subsections, we isolated each adversarial perturbation type into a separate evaluation set so that researchers can diagnose the robustness of their developed models from different aspects. This will help researchers understand which aspects of robustness require further enhancement, and improve their models accordingly. 
We also added a mix-perturbation evaluation set by combining two or three different-level annotated perturbations for each example. This evaluation set provides insights about the overall robustness of Table QA models.

\section{Diagnostic Experiments}
In this section, we evaluate existing Table QA models on our constructed benchmark, \ours.

\subsection{Experimental Setup}

\paragraph{Compared Table QA models}

We evaluated the following four representative table QA models on \ours, which first pre-trained on the collected large table corpus and then fine-tuned on the downstream Table QA tasks.

\begin{itemize} [leftmargin=*]
   \item \textbf{\tapas}~\cite{herzig-etal-2020-tapas} is based on BERT’s encoder with additional positional embeddings for encoding tabular structure and two classification layers for cell selection and aggregation operator predictions. 
   \item \textbf{TableFormer}~\cite{yang-etal-2022-tableformer} adapts \tapas by introducing a learnable attention biases to mitigate the unwanted bias brought from row and column encoding. 
   \item \textbf{TAPEX}~\cite{liu2022tapex} models the Table QA as a sequence-to-sequence task, and uses BART~\cite{lewis-etal-2020-bart} as the backbone without any table-relevant architecture design. 
   \item \textbf{OmniTab}~\cite{jiang-etal-2022-omnitab} uses the same backbone as TAPEX, and is further pre-trained on collected natural and synthetic Table QA examples. 
\end{itemize}
We also evaluated the \textbf{GPT-3}~\cite{gpt-3} model in a few-shot setting.

\paragraph{Implementation Details}
Since \ours only includes evaluation data, we fine-tuned the \texttt{Large} version of each Table QA model using the original Table QA training set and obtained three variants for WTQ, \wikisql, and SQA. As WTQ and SQA datasets have multiple official train/dev splits for the purpose of cross-validation, we used the split of \texttt{random-split-1-train} for fine-tuning. Specifically, WTQ training set contains 11,321 examples, \wikisql training set contains 56,355 examples, and SQA training set contains 4,257 sequences. We randomly split each official training set into a train/dev set with a ratio of 8:2 for fine-tuning. 
We ran 20 epochs with a batch size of 128 for each fine-tuning experiments and selected the best fine-tuning checkpoint based on the validation loss on the splitted dev set.

In terms of GPT-3 few-shot experiments, we used \texttt{text-davinci-003} via the public OpenAI APIs\footnote{\url{https://openai.com/api/}} with \emph{two-shot} prompting. Similar to \citet{table-cot}, we used a temperature of 0.7 without any frequency penalty and without top-k truncation. An example of ``chain-of-thought” prompt prefix is shown in Figure~\ref{fig:GPT3qa} in Appendix.

\paragraph{Evaluation Metrics}
We used \emph{Exact Match Accuracy} as the evaluation metric, which checks whether the predicted answers are equal to the ground truth. For SQA, we reported the average  accuracy for sequential questions. We used the following three metrics to evaluate model robustness: \textbf{Pre-perturbation Accuracy} over pre-perturbation data; \textbf{Post-perturbation Accuracy} over post-perturbation data; 
\textbf{Robustness Accuracy} as the ratio of the correct predictions on both pre- and post-perturbation data versus the correct predictions on pre-perturbation data.

\subsection{Diagnostic Results}
According to Table~\ref{tab:wtq_attack_results} and Table~\ref{tab:wikisql_results}, \ref{tab:sqa_results} in Appendix, all examined Table QA models exhibited significant performance drops for each perturbation type, thus are not robust under adversarial attacks.

\paragraph{Effect of Model Architecture} 
We found that TableFormer is the most robust against row and column order shuffling, with the help of its task-independent relative attention mechanism for table-text encoding. Despite that, for most perturbation types, \tapas and TableFormer, even with specific table encoding designs, do not outperform TAPEX and OmniTab in robustness. Therefore, we conclude that model architectures may help defend specific but not all perturbation attacks.

\begin{table*}[!t]
\centering
\small
\begin{tabular}{llcccccccc}
\toprule
\multirow{2}{*}{\begin{tabular}[c]{@{}l@{}}Level\end{tabular}} & \multirow{2}{*}{\begin{tabular}[c]{@{}l@{}}Perturbation Type\end{tabular}} & \multicolumn{2}{c}{\texttt{text-davinci-002}} && \multicolumn{2}{c}{\texttt{text-davinci-003}} && \multicolumn{2}{c}{\texttt{gpt-3.5-turbo}}\\ 
\noalign{\vskip 1ex}\cline{3-4} \cline{6-7} \cline{9-10} \noalign{\vskip 1ex}

 & & \textsc{Post-Acc}& \textsc{R-Acc} && \textsc{Post-Acc}& \textsc{R-Acc} &&\textsc{Post-Acc}& \textsc{R-Acc}\\
\midrule
 &  Development Set & 40.3 & -- && 42.9 & -- && 43.7 & --\\
\noalign{\vskip 1ex}\cdashline{1-10}\noalign{\vskip 1ex}
 
\multirow{2}{*}{\begin{tabular}[c]{@{}l@{}}Table\\Header\end{tabular}} 
& Synonym Replace & 39.2 & 87.8 && 39.9 & 90.7 && \bf{42.0} & \textbf{90.9}\\
& Abbrev Replace &  37.1 & 90.1 && 39.2 & 93.8 && \bf{40.0} & \bf{94.4}\\

\noalign{\vskip 1ex}\cdashline{1-10}\noalign{\vskip 1ex}

\multirow{5}{*}{\begin{tabular}[c]{@{}l@{}}Table\\Content\end{tabular}} 
& Row Shuffle & 35.7 & 87.4 && \bf{38.5} & \bf{90.2} && 36.6 & 88.7\\
& Col Shuffle & 36.0 & 90.4 && \bf{40.0} & \bf{93.3} && 39.5 & 92.6\\
& Col Extension & 35.2 & 79.5 && 37.4 & 81.4 && \bf{38.0} & \bf{82.0}\\
& Col Mask & 40.1 & 94.0 && 41.9 & \bf{97.0} && \bf{42.2} & 96.4\\
& Col Add & 33.7 & 80.2 && 36.8 & 85.6 && \bf{37.0} & \bf{86.1}\\
\noalign{\vskip 1ex}\cdashline{1-10}\noalign{\vskip 1ex}

\multirow{2}{*}{\begin{tabular}[c]{@{}l@{}}NLQ\end{tabular}} 
& Word-Level & 37.9 & 93.3 && 40.3 & \bf{93.7} && \bf{40.7} & 93.5\\
& Sentence-Level & 40.6 & 93.7 && 40.5 & \bf{94.2} && \bf{41.2} & 93.4\\
\noalign{\vskip 1ex}\cdashline{1-10}\noalign{\vskip 1ex}

Mix &--& 29.7 & 82.5 && 30.6 & 83.2 && \bf{31.4} & \bf{84.9}\\
\bottomrule
\end{tabular}

\caption{\emph{Post-perturbation Accuracy} and \emph{Robustness Accuracy} of GPT series models on \ours-WTQ. Models with higher post-perturbation accuracy correlated with higher robustness accuracy in most cases. Due to the budget constraints, we reported results on 200 randomly
sampled examples for each perturbation type.}
\label{tab:gpt_wtq}
\end{table*}

\paragraph{Large Language Model is more Robust} In context learning with GPT-3 is more robust than other models in most perturbation categories. First, the significantly larger pre-training corpus size and model parameters allow GPT-3 to better generalize to new data~\cite{wei2022emergent}. Second, as discussed in Section~\ref{sec:table_content_perturb}, existing Table QA datasets contain \emph{annotation bias} related to both table contents and questions. And fine-tuned models, therefore, learn shortcuts to overfit to the training data, which limits their ability to defend against perturbations. To provide more insights into the robustness of in-context learning with large language models, we also evaluated various types of GPT series models (i.e., \texttt{text-davinci-002}, \texttt{text-davinci-003}, and \texttt{gpt-3.5-turbo}) on the \ours-WTQ set. As shown in Table~\ref{tab:gpt_wtq}, GPT series models with higher post-perturbation accuracy correlated with higher robustness accuracy in most cases.

\section{\framework Framework}
Motivated by the diagnostic results that LLMs are more robust against human-annotated perturbations, we adopted LLMs to enhance the robustness of \emph{smaller} (i.e., less than 1B parameter) and \emph{fine-tuned} Table QA models.
Specifically, we introduced \textbf{L}LM-\textbf{\textsc{e}}nhanced \textbf{T}able QA \textbf{A}ugmentation (\framework) framework, which generates adversarial training examples at scale using the LLM prompting method, to improve model robustness against human-annotated adversarial perturbations. 

Specifically, for each perturbation type in \ours,  we designed task-specific ``chain-of-thought” prompts~\cite{cot, table-cot} to guide the GPT-3 (i.e., \texttt{text-davinci-003}) or CodeX (i.e., \texttt{code-davinci-002}) models to generate adversarial examples to enhance the training set. We repeated example generation three times to create diverse training data. We next discuss the details for each augmentation level.

\subsection{Table Header Augmentation}
For both \emph{header synonym} and \emph{header abbreviation replacements} type, we randomly selected 10 examples from human-annotated perturbations as demonstrations. Each example includes the table header and first two rows as input and the perturbed table header as output (Figure~\ref{fig:header_prompt} in Appendix).

\subsection{Table Content Augmentation}
For \emph{column extension} and \emph{column masking} types, we provided 8  demonstration examples. Each example includes the original table, the extended (or masked) column, and the corresponding explanations. 
For \emph{column adding} type, we applied an existing table dense retriever to find the three most relevant tables (Section~\ref{sec:dense_retrieve}), and then prompted the CodeX model to added one or two new columns from the retrieved tables. Figure~\ref{fig:table_prompt} in Appendix shows a prompt prefix example for \emph{column adding}.
For \emph{row} or \emph{column order shuffling}, we used heuristics to produce perturbed source table variants.

\subsection{NLQ Augmentation}
We analyzed the human-annotated perturbed questions and summarized three paraphrase categories at the word level, and two categories at the sentence level. Table~\ref{tab:nlq_example} in Appendix shows examples for each category.

\paragraph{Word-level NLQ} We focused on paraphrasing three types of question words: 
1) \emph{reasoning operation indicators} (e.g., ``how many'' - counting operation), to infer the reasoning type; 
2) \emph{table header indicators} (e.g., ``who" - ``athlete'' column), to locate the relevant columns; 
and 3) \emph{cell value indicators} (e.g., US - ``USA'' cell), to locate the relevant cells.

\paragraph{Sentence-level NLQ} 
We designed two task-specific perturbations in terms of \emph{sentence simplification} (e.g., ``at the first place of'' -  ``number one'')  and \emph{interrogative transformation} (e.g., ``when was'' - ``Please provide me with''). We also included \emph{general syntactic perturbations} (e.g., ``stock codes'' - ``ticker symbols'') in sentence-level paraphrasing.

For each paraphrase category at word and sentence level, we designed five to eight demonstration examples to prompt GPT for paraphrased questions. Each example includes the original question, paraphrased question, and corresponding explanations.

\begin{table*}[!t]
\centering
\small
\begin{tabular}{llcccccc}
\toprule
\multirow{2}{*}{Level} & \multirow{2}{*}{Perturbation Type} & \multirow{2}{*}{TAPAS} & &  & \multirow{2}{*}{TAPEX} &  & \\
& & & w/ RTA & w/ \framework & & w/ RTA & w/ \framework \\
\midrule
 &  Development Set & \textbf{48.3} & 45.3 (- 3.0) & 46.5 (- 1.8) & \textbf{57.3} & 53.6 (- 3.7) & 55.3 (- 2.0)\\
\noalign{\vskip 1ex}\cdashline{1-8}\noalign{\vskip 1ex}
 
\multirow{2}{*}{\begin{tabular}[c]{@{}l@{}}Table\\Header\end{tabular}} 
& Synonym Replace & 38.5 & 40.8 (+2.3) & \textbf{42.4} (+3.9) & 48.4 & 51.0 (+2.6) &  \textbf{52.5} (+4.1)\\
& Abbrev Replace &  35.1 & 38.9 (+3.8) & \textbf{40.7} (+5.6) & 44.3 & 48.7 (+4.4) &  \textbf{50.0} (+5.7)\\

\noalign{\vskip 1ex}\cdashline{1-8}\noalign{\vskip 1ex}

\multirow{5}{*}{\begin{tabular}[c]{@{}l@{}}Table\\Content\end{tabular}} 
& Row Shuffle & 40.6 & \textbf{42.3} (+1.7) & 42.2 (+1.6) & 45.7 & 48.1 (+2.4) &  \textbf{48.2} (+2.5)\\
& Col Shuffle & 42.5 & \textbf{43.8} (+1.3) & 43.6 (+1.1) & 48.5 &  \textbf{50.1} (+1.6) &  \textbf{50.1} (+1.6)\\
& Col Extension & 42.5 & 44.2 (+1.7) & \textbf{46.3} (+3.8) & 47.8 & 50.0 (+2.2) & \textbf{51.3} (+3.5)\\
& Col Mask & 45.2 & 45.4 (+0.2) & \textbf{45.6} (+0.4) & 54.4 & 54.3 (- 0.1) & \textbf{54.6} (+0.2)\\
& Col Add & 47.1 & 47.6 (+0.5) & \textbf{47.9} (+0.8) & 50.4 & 53.1 (+2.7) & \textbf{54.2} (+3.8)\\
\noalign{\vskip 1ex}\cdashline{1-8}\noalign{\vskip 1ex}

\multirow{2}{*}{\begin{tabular}[c]{@{}l@{}}NLQ\end{tabular}} 
& Word-Level & 38.6 & 41.0 (+2.4) & \textbf{43.1} (+4.5) & 49.2 & 51.0 (+1.8) & \textbf{52.4} (+3.2)\\
& Sentence-Level & 41.1 & 41.7 (+0.6) & \textbf{43.6} (+2.5) & 49.5 & 50.7 (+1.2) & \textbf{52.9} (+3.4)\\
\noalign{\vskip 1ex}\cdashline{1-8}\noalign{\vskip 1ex}

Mix &--& 32.0 & 33.1 (+1.1) & \textbf{35.2} (+3.2) & 39.5 & 41.0 (+1.5) & \textbf{42.3} (+2.8)\\
\bottomrule
\end{tabular}
\caption{Accuracy of \tapas and TAPEX models on \ours-WTQ before and after adversarial training. Compared with RTA, \framework-augmented models have higher accuracy improvement across most types of perturbations.}
\label{tab:defense_results}
\end{table*}

\begin{table*}[!t]
\centering
\begin{ADLinactivate}
\small
\begin{tabular}{ll|cc|cc|cc}
\toprule
\multirow{2}{*}{Level} & \multirow{2}{*}{Type} & \multicolumn{2}{c|}{\%S $\geq$ 4} & \multicolumn{2}{c|}{\begin{tabular}[c]{@{}c@{}}\% win\end{tabular}} & \multicolumn{2}{c}{\begin{tabular}[c]{@{}c@{}}$\approx$ \$ Cost (100 examples)\end{tabular}}\\ 
\cmidrule(lr){3-4} \cmidrule(lr){5-6} \cmidrule(lr){7-8} 
& & Human & \framework & Human & \framework & Human & \framework \\
\midrule
\multirow{2}{*}{\begin{tabular}[c]{@{}l@{}}Table\\Header\end{tabular}} 
& Synonym & 95.5 & 90.0 & 69 & 52 & 60.0 & 1.5\\
& Abbreviation & 90.5 & 82.5 & 76 & 41 & 60.0 & 1.5\\\midrule
\multirow{3}{*}{\begin{tabular}[c]{@{}l@{}}Table\\Content\end{tabular}} 
& Col Extend & 90.0 & 63.5 & 90 & 22 & 100.0 & 6.0\\
& Col Mask & 91.5 & 69.0 & 85 & 27 & 60.0 & 6.0\\
& Col Add & 92.0 & 70.0 & 83 & 35 & 30.0 & 8.5\\\midrule
\multirow{2}{*}{NLQ} 
& Word-level & 96.0 & 90.0 & 70 & 56 & 80.0 & 1.5\\
& Sent-level & 94.0 & 92.0 & 74 & 50 & 80.0 & 1.5\\
\bottomrule
\end{tabular}
\end{ADLinactivate}
\caption{Comparison of adversarial data quality and cost of human annotation and \framework. We report 1) percent of samples that have an average score $\geq$ 4, and 2) Percentage of times the examples are selected as better (may be tied). \framework achieves comparable performance to human annotators for table header and NLQ perturbations, with a significantly lower annotation cost. We regard the pricing for \texttt{code-davinci-002} used in table content augumentation the same as \texttt{text-davinci-003} (i.e., \$0.02/1K tokens).}
\label{tab:annotation_comparsion}
\end{table*}
\begin{table}[t!]
\centering
\small
\begin{tabular}{p{1.25cm}p{5.5cm}}
\toprule
Error Type & Example \\
\midrule
Change original meaning & 
\textbf{Original}: How many districts were created in the 1900’s?\newline
\textbf{Paraphrased}: How many districts were created in the \red{nineteenth} century?\newline
\textbf{Explanation}: Should be \textit{\red{twentieth}}
\\\noalign{\vskip 0.5ex}\cdashline{1-2}\noalign{\vskip 0.5ex}

Mismatch with given prompt & For the prompt of replacing carrier phrase\newline
\textbf{Original}: How many players scored more than 7 points?\newline
\textbf{Paraphrased}: How many \red{athletes} scored more than 7 points?\newline
\textbf{Explanation}: Should paraphrase the carrier phrase \textit{\red{How many}} 
\\\noalign{\vskip 0.5ex}\cdashline{1-2}\noalign{\vskip 0.5ex}

Information missing & \textbf{Original}: What are the names and \red{stock code} of companies whose headquarters are located in the United States?\newline
\textbf{Paraphrased}: Name some companies whose headquarters are located in the United States.\newline
\textbf{Explanation}: \textit{\red{stock code}} is missing
\\\noalign{\vskip 0.5ex}\cdashline{1-2}\noalign{\vskip 0.5ex}
Hallucination & \textbf{Original}: What is the name and nation of the singer who have a song having ``Hey'' in its name?\newline
\textbf{Paraphrased}: What is the name and nation of the singer having a song named \red{``Hey Ya!”} \newline
\textbf{Explanation}: \textit{\red{``Hey Ya!”}} does not appear in the given context \\

\bottomrule
\end{tabular}
\caption{Case study for common errors made by the \framework framework for NLQ perturbation. The colored text highlights model errors.}
\label{tab:error_cases}
\end{table}
\section{Adversarial Training Experiments}
In this section, we evaluate \framework on our constructed benchmarks, \ours.

\subsection{Experiment Setup}
\paragraph{Baseline System} To compare with \framework, we developed a competitive adversarial training data generation pipeline, RTA, which applied rule-based methods to generate adversarial augmentation data for each perturbation type in terms of table header and table content. It further used BERT-Attack~\cite{li-etal-2020-bert-attack} to generate paraphrased questions. 

\paragraph{Implementation Details}
We selected \tapas and TAPEX for experiments because they are the foundations of TableFormer and OmniTab, respectively. We evaluated the model performance on \ours-WTQ before and after adversarial training. Models were fine-tuned from scratch on corresponding augmented training sets, which included both original and adversarial training data.

\subsection{Results}
According to Table~\ref{tab:defense_results}, compared with RTA, \framework-augmented models have higher post-perturbation accuracy across most types of \ours perturbations. This result demonstrates that using LLM-prompting methods to generate adversarial training examples is more effective. 
In addition, despite the model's performance on the original development set decreasing with augmented data, the \framework-augmented models are better on the original development set than the RTA-augmented models. This suggests that \framework introduces less noise into the original training sets, as LLMs generate more natural adversarial examples. 
Such trade-off between robustness and accuracy (i.e., adversarial robustness comes at the cost of standard performance) has also been widely observed and discussed in different ML/NLP areas~\cite{tsipras2018robustness, pmlr-v97-zhang19p, zhao-etal-2022-bridging}. We will explore how to improve robustness without compromising accuracy in our future work.

\subsection{Analysis}
To evaluate the quality of adversarial example generation, we conducted human evaluations to compare the quality of the examples generated by the \framework framework with those created by human annotators. We further provided case studies on common errors made by \framework.

\paragraph{Comparison with Human Annotation}
For each perturbation type, we sampled 100 adversarial examples from both human annotation and \framework. Two evaluators were then asked to rate each sample on a scale of 1 (worst) to 5 (best) and determine which example was better, between the one created by human annotators and the framework. We also estimated the annotation cost of each perturbation type for both methods. The results in Table~\ref{tab:annotation_comparsion} demonstrate that \framework achieves comparable performance to human annotators for table header and NLQ perturbations, with much lower annotation cost. However, it still lags behind human annotators in terms of table content perturbations, we leave future work to design more effective prompting methods for table content augmentation.

\paragraph{Error analysis of \framework generation}
Table~\ref{tab:error_cases} shows examples of perturbed questions generated by the \framework framework. We identified the following common mistakes that \framework are likely to make: 1) changing the original meaning of the questions; 2) not consistent with the demonstration in the given prompt; 3) missing important information from the original question; and 4) hallucination.

\section{Conclusion}
This work proposes \ours, the first benchmark for \tableqa robustness. \ours measures the robustness of Table QA models against different levels of human-annotated perturbations. Experimental results showed that state-of-the-art models exhibited significant performance drops on our \ours benchmark. To address this issue, we designed the \framework framework, which utilizes LLM-promoting methods to generate adversarial training examples to enhance Table QA model robustness. We believe that our work will raise awareness among researchers about the importance of robustness in Table QA models.
\section*{Acknowledgements}
We would like to dedicate this paper to the memory of Dr. Dragomir Radev. Dr. Radev's leadership, guidance, and expertise were instrumental in shaping the direction and quality of this project. His loss is deeply felt by all of us involved. We extend our heartfelt gratitude to Dr. Radev for his passion and dedication to the NLP community.

Chen Zhao is supported by the DARPA PTG program.  Any opinions, findings, and conclusions or recommendations expressed in this material are those of the author(s) and do not necessarily reflect the views of the DARPA.

\section*{Limitations}
This work focuses on diagnosing and enhancing model robustness for Table QA tasks. However, there are other types of table reasoning benchmarks, such as table fact checking~\cite{Chen2020TabFact, gupta-etal-2020-infotabs, aly2021feverous} and logical table-to-text generation~\cite{chen-etal-2020-logical, hitab}, whose model robustness has not been well explored. We believe future work could extend the approaches for constructing \ours to these other table reasoning benchmarks, providing a more comprehensive understanding of model robustness for table understanding and reasoning tasks. Moreover, we did not consider those perturbations related to modifying the original cell values, which might change the final answer and thus will take a longer time for annotation. We believe future work could explore perturbations at the cell level.

\section*{Ethical Consideration}
\ours were constructed upon the development set of WTQ~\cite{pasupat-liang-2015-compositional}, \wikisql~\cite{zhongSeq2SQL2017}, and SQA~\cite{iyyer-etal-2017-search} datasets, which are publicly available under the licenses of CC BY-SA 4.0\footnote{\url{https://creativecommons.org/licenses/by-sa/4.0/}}, BSD 3-Clause\footnote{\url{https://opensource.org/licenses/BSD-3-Clause}}, and MIT\footnote{\url{https://opensource.org/licenses/MIT}}, respectively. These licenses all permit us to compose, modify, publish, and distribute additional annotations upon the original dataset. All the experiments in this paper can be run on a single NVIDIA Tesla V100-32G GPU. Our benchmark and code will be released along with the paper.

For the \ours annotation, we hired 15 graduate students (9 females and 6 males) majoring in STEM majors. The hourly rates are in the range of \$10 and \$12 based on the different working speed (above the local average wage of similar jobs). We recommended that annotators spend at most 4 hours per day for annotation in order to reduce pressure and maintain a comfortable pace. The whole annotation work lasted about 30 days.

\bibliography{custom}
\bibliographystyle{acl_natbib}

\appendix
\section{Appendix} 
\begin{figure}[H]
    \centering
    \includegraphics[width = \linewidth]{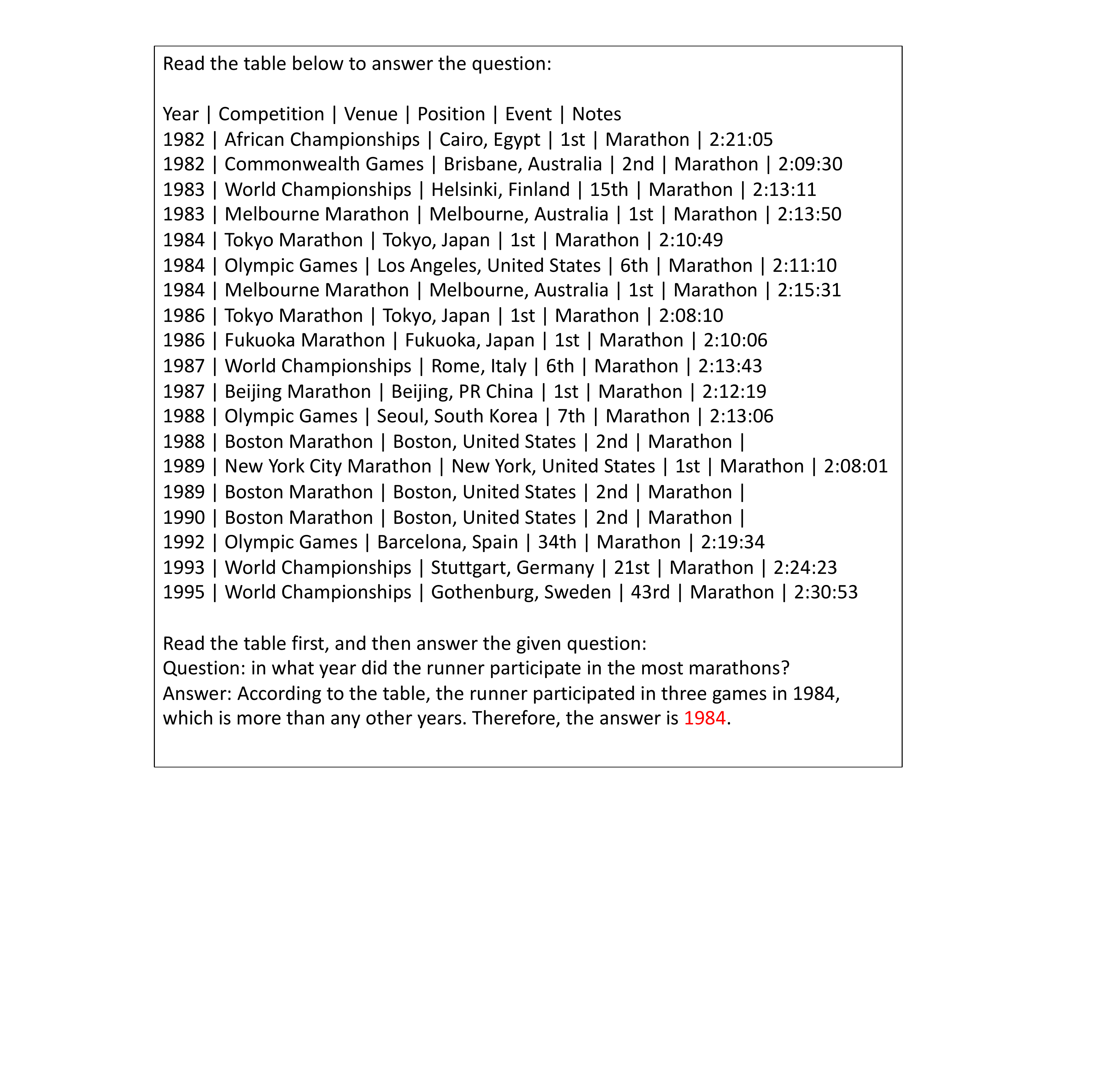}
    \caption{An example of GPT-3 ``chain-of-thought” prompt prefix for the Table QA tasks.}
    \label{fig:GPT3qa}
\end{figure}
\begin{table*}[!t]
\setlength\tabcolsep{3pt}
\centering
\resizebox{0.98\textwidth}{!}{
\begin{tabular}{llrc*{4}{ccc}cc}
\toprule
\multirow{2}{*}{\begin{tabular}[c]{@{}l@{}}Level\end{tabular}} & \multirow{2}{*}{\begin{tabular}[c]{@{}l@{}}Perturbation Type\end{tabular}} & \multirow{2}{*}{\begin{tabular}[c]{@{}l@{}}\# Example\end{tabular}} && \multicolumn{2}{c}{TAPAS} && \multicolumn{2}{c}{TableFormer} && \multicolumn{2}{c}{TAPEX} && \multicolumn{2}{c}{OmniTab} && \multicolumn{2}{c}{GPT-3} \\ 
\noalign{\vskip 0.5ex}\cline{5-6} \cline{8-9} \cline{11-12} \cline{14-15} \cline{17-18}\noalign{\vskip 0.5ex}

 & & & &\textsc{Acc}& \textsc{R-Acc} && \textsc{Acc}& \textsc{R-Acc} &&\textsc{Acc}& \textsc{R-Acc}&&\textsc{Acc}& \textsc{R-Acc} &&\textsc{Acc}& \textsc{R-Acc} \\
\midrule
 & Development Set & 8,421 && 87.1 & -- && 85.8 & -- && 89.5 & -- && 88.8 & -- && 78.3 & -- \\
\midrule
 
\multirow{4}{*}{\begin{tabular}[c]{@{}l@{}}Table\\Header\end{tabular}} 
& \multirow{2}{*}{\begin{tabular}[c]{@{}l@{}}Synonym Replacement\end{tabular}} &
\multirow{2}{*}{9,419} && 81.2 / 62.2 & \multirow{2}{*}{73.1} & & 80.7 / 64.0 & \multirow{2}{*}{75.8} & & 83.6 / 68.8 & \multirow{2}{*}{79.5} & & 82.3/70.7 & \multirow{2}{*}{\underline{82.0}} & & 78.1 / 74.5 & \multirow{2}{*}{\textbf{91.8}} \\
& & & & \color{red}{(-19.0)} && & \color{red}{(-16.7)} && & \color{red}{(-14.8)} && & \color{red}{(-11.6)} && & \color{red}{(-3.6)} &\\
\noalign{\vskip 0.5ex}\cdashline{2-18}\noalign{\vskip 0.5ex}

& \multirow{2}{*}{\begin{tabular}[c]{@{}l@{}}Abbreviation Replacement\end{tabular}} &
\multirow{2}{*}{8,229} && 81.7 / 59.5 & \multirow{2}{*}{69.9} & & 81.0 / 57.7 & \multirow{2}{*}{66.7} & & 82.9/70.7 & \multirow{2}{*}{82.5} & & 82.1 / 73.2 & \multirow{2}{*}{\underline{85.8}} & & 78.5 / 75.1 & \multirow{2}{*}{\textbf{89.1}} \\
& & & & \color{red}{(-22.2)} && & \color{red}{(-23.3)} && & \color{red}{(-22.2)} && & \color{red}{(-18.9)} && & \color{red}{(-3.4)} &\\
\midrule

\multirow{10}{*}{\begin{tabular}[c]{@{}l@{}}Table\\Content\end{tabular}} 
& \multirow{2}{*}{\begin{tabular}[c]{@{}l@{}}Row Order Shuffling\end{tabular}} &
\multirow{2}{*}{17,490} && 84.8 / 80.1 & \multirow{2}{*}{91.1} & & 85.7 / 85.2 & \multirow{2}{*}{\textbf{96.9}} & & 88.5 / 83.0 & \multirow{2}{*}{86.2} & & 87.6 / 82.4 & \multirow{2}{*}{87.9} & & 78.2 / 76.5 & \multirow{2}{*}{\underline{92.3}} \\
& & & & \color{red}{(-4.7)} && & \color{red}{(-0.5)} && & \color{red}{(-5.5)} && & \color{red}{(-5.2)} && & \color{red}{(-1.7)} &\\
\noalign{\vskip 0.5ex}\cdashline{2-18}\noalign{\vskip 0.5ex}

& \multirow{2}{*}{\begin{tabular}[c]{@{}l@{}}Column Order Shuffling\end{tabular}} &
\multirow{2}{*}{16,532} && 85.6 / 83.9 & \multirow{2}{*}{93.0} & & 84.9 / 84.8 & \multirow{2}{*}{\textbf{99.3}} & & 89.0 / 87.4 & \multirow{2}{*}{92.1} & & 87.6 / 85.3 & \multirow{2}{*}{90.4} & & 77.5 / 76.9 & \multirow{2}{*}{\underline{94.4}} \\
& & & & \color{red}{(-1.7)} && & \color{red}{(-0.1)} && & \color{red}{(-1.6)} && & \color{red}{(-2.3)} && & \color{red}{(-0.6)} &\\
\noalign{\vskip 0.5ex}\cdashline{2-18}\noalign{\vskip 0.5ex}

& \multirow{2}{*}{\begin{tabular}[c]{@{}l@{}}Column Extension\end{tabular}} &
\multirow{2}{*}{2,626} && 89.8 / 51.9 & \multirow{2}{*}{56.4} & & 86.0 / 50.8 & \multirow{2}{*}{\textbf{58.8}} & & 92.0 / 53.2 & \multirow{2}{*}{57.1} & & 91.2 / 53.8 & \multirow{2}{*}{\underline{57.6}} & & 80.2 / 55.5 & \multirow{2}{*}{56.4} \\
& & & & \color{red}{(-37.9)} && & \color{red}{(-35.2)} && & \color{red}{(-38.8)} && & \color{red}{(-37.4)} && & \color{red}{(-34.7)} &\\
\noalign{\vskip 0.5ex}\cdashline{2-18}\noalign{\vskip 0.5ex}

& \multirow{2}{*}{\begin{tabular}[c]{@{}l@{}}Column Masking\end{tabular}} &
\multirow{2}{*}{1,153} && 85.1 / 79.2 & \multirow{2}{*}{\textbf{87.4}} & & 84.8 / 76.9 & \multirow{2}{*}{85.0} & & 89.5 / 82.4 & \multirow{2}{*}{80.6} & & 88.6 / 82.1 & \multirow{2}{*}{81.5} & & 78.2 / 74.7 & \multirow{2}{*}{\underline{85.6}} \\
& & & & \color{red}{(-5.9)} && & \color{red}{(-7.9)} && & \color{red}{(-7.1)} && & \color{red}{(-6.5)} && & \color{red}{(-3.5)} &\\
\noalign{\vskip 0.5ex}\cdashline{2-18}\noalign{\vskip 0.5ex}

& \multirow{2}{*}{\begin{tabular}[c]{@{}l@{}}Column Adding\end{tabular}} &
\multirow{2}{*}{6,444} && 77.8 / 69.6 & \multirow{2}{*}{\textbf{81.6}} & & 75.4 / 67.3 & \multirow{2}{*}{80.1} & & 81.4 / 64.9 & \multirow{2}{*}{71.0} & & 79.7 / 66.7 & \multirow{2}{*}{71.3} & & 78.3 / 70.5 & \multirow{2}{*}{\underline{81.0}} \\
& & & & \color{red}{(-8.2)} && & \color{red}{(-8.1)} && & \color{red}{(-16.5)} && & \color{red}{(-13.0)} && & \color{red}{(-7.8)} &\\

\midrule

\multirow{4}{*}{\begin{tabular}[c]{@{}l@{}}NLQ\end{tabular}} 
& \multirow{2}{*}{\begin{tabular}[c]{@{}l@{}}Word-Level Paraphrase\end{tabular}} &
\multirow{2}{*}{5,024} && 82.7 / 58.9 & \multirow{2}{*}{68.0} & & 82.1 / 57.0 & \multirow{2}{*}{66.7} & & 85.8 / 64.2 & \multirow{2}{*}{\underline{72.6}} & & 84.7 / 64.3 & \multirow{2}{*}{\underline{72.6}} & & 76.3 / 72.2 & \multirow{2}{*}{\textbf{92.2}} \\
& & & & \color{red}{(-23.8)} && & \color{red}{(-25.1)} && & \color{red}{(-21.6)} && & \color{red}{(-20.4)} && & \color{red}{(-4.1)} &\\
\noalign{\vskip 0.5ex}\cdashline{2-18}\noalign{\vskip 0.5ex}

& \multirow{2}{*}{\begin{tabular}[c]{@{}l@{}}Sentence-Level Paraphrase\end{tabular}} &
\multirow{2}{*}{3,726} && 79.3 / 66.7 & \multirow{2}{*}{78.6} & & 76.8 / 64.5 & \multirow{2}{*}{79.1} & & 81.6 / 68.7 & \multirow{2}{*}{\underline{80.8}} & & 80.6 / 70.1 & \multirow{2}{*}{81.3} & & 75.0 / 72.6 & \multirow{2}{*}{\textbf{95.1}} \\
& & & & \color{red}{(-12.6)} && & \color{red}{(-12.3)} && & \color{red}{(-12.9)} && & \color{red}{(-10.5)} && & \color{red}{(-2.4)} &\\\midrule

\multirow{2}{*}{Mix} 
& \multirow{2}{*}{--}  &
\multirow{2}{*}{4,752} && 70.8 / 52.9 & \multirow{2}{*}{69.9} & & 70.1 / 51.2 & \multirow{2}{*}{67.5} & & 80.1 / 60.3 & \multirow{2}{*}{70.7} & & 79.2 / 64.2 & \multirow{2}{*}{\underline{71.2}} & & 69.5 / 60.1 & \multirow{2}{*}{\textbf{80.2}} \\
& & & & \color{red}{(-17.9)} && & \color{red}{(-18.9)} && & \color{red}{(-19.8)} && & \color{red}{(-18.0)} && & \color{red}{(-9.4)} &\\

\bottomrule
\end{tabular}
}
\caption{Data statistics and robustness evaluation results of state-of-the-art Table QA models on \ours-\wikisqlshort. \textsc{Acc} represents the \emph{Pre-} and \emph{Post-perturbation Accuracy}; \textsc{R-Acc} represents the \emph{Robustness Accuracy}. Bold numbers indicate the highest \emph{Robustness Accuracy} in each perturbation type, and underscores denote the second best result. When evaluating GPT-3 in a few-shot setting, we reported results on 200 randomly sampled examples for each perturbation type.}
\label{tab:wikisql_results}
\end{table*}
\begin{table*}[!t]
\setlength\tabcolsep{3pt}
\centering
\resizebox{0.75\textwidth}{!}{%
\begin{tabular}{llrc*{4}{ccc}cc}
\toprule
\multirow{2}{*}{\begin{tabular}[c]{@{}l@{}}Level\end{tabular}} & \multirow{2}{*}{\begin{tabular}[c]{@{}l@{}}Perturbation Type\end{tabular}} & \multirow{2}{*}{\begin{tabular}[c]{@{}l@{}}\# Example\end{tabular}} && \multicolumn{2}{c}{TAPAS} && \multicolumn{2}{c}{TAPEX} && \multicolumn{2}{c}{GPT-3} \\ 
\noalign{\vskip 0.5ex}\cline{5-6} \cline{8-9} \cline{11-12} \noalign{\vskip 0.5ex}

 & & & &\textsc{Acc}& \textsc{R-Acc} && \textsc{Acc}& \textsc{R-Acc} &&\textsc{Acc}& \textsc{R-Acc} \\
\midrule
 & Development Set & 784 && 63.7 & -- && 67.9 & -- && 50.1 & --  \\
\midrule
 
\multirow{4}{*}{\begin{tabular}[c]{@{}l@{}}Table\\Header\end{tabular}} 
& \multirow{2}{*}{\begin{tabular}[c]{@{}l@{}}Synonym Replacement\end{tabular}} &
\multirow{2}{*}{2,104} && 64.4 / 57.7 & \multirow{2}{*}{85.7} & & 68.6 / 62.0 & \multirow{2}{*}{86.5} & & 50.7 / 47.2 & \multirow{2}{*}{\textbf{91.3}} \\
& & & & \color{red}{(-6.7)} && & \color{red}{(-6.6)} && & \color{red}{(-3.5)} &\\
\noalign{\vskip 0.5ex}\cdashline{2-12}\noalign{\vskip 0.5ex}

& \multirow{2}{*}{\begin{tabular}[c]{@{}l@{}}Abbreviation Replacement\end{tabular}} &
\multirow{2}{*}{1,286} && 62.9 / 50.0 & \multirow{2}{*}{76.9} & & 68.5 / 59.7 & \multirow{2}{*}{83.8} & & 51.0 / 47.3 & \multirow{2}{*}{\textbf{90.6}} \\
& & & & \color{red}{(-12.9)} && & \color{red}{(-8.8)} && & \color{red}{(-3.7)} &\\
\midrule

\multirow{10}{*}{\begin{tabular}[c]{@{}l@{}}Table\\Content\end{tabular}} 
& \multirow{2}{*}{\begin{tabular}[c]{@{}l@{}}Row Order Shuffling\end{tabular}} &
\multirow{2}{*}{2,356} && 60.9 / 55.3 & \multirow{2}{*}{85.0} & & 64.1 / 60.2 & \multirow{2}{*}{88.9} & & 49.2 / 47.4 & \multirow{2}{*}{\textbf{93.7}} \\
& & & & \color{red}{(-5.6)} && & \color{red}{(-3.9)} && & \color{red}{(-1.8)} &\\
\noalign{\vskip 0.5ex}\cdashline{2-12}\noalign{\vskip 0.5ex}

& \multirow{2}{*}{\begin{tabular}[c]{@{}l@{}}Column Order Shuffling\end{tabular}} &
\multirow{2}{*}{2,079} && 61.3 / 60.5 & \multirow{2}{*}{\textbf{94.8}} & & 66.7 / 65.2 & \multirow{2}{*}{89.8} & & 49.5 / 49.0 & \multirow{2}{*}{94.5} \\
& & & & \color{red}{(-0.8)} && & \color{red}{(-1.5)} && & \color{red}{(-0.5)} &\\
\noalign{\vskip 0.5ex}\cdashline{2-12}\noalign{\vskip 0.5ex}

& \multirow{2}{*}{\begin{tabular}[c]{@{}l@{}}Column Extension\end{tabular}} &
\multirow{2}{*}{1,540} && 62.4 / 40.8 & \multirow{2}{*}{62.1} & & 66.8 / 42.0 & \multirow{2}{*}{58.9} & & 49.5 / 34.9 & \multirow{2}{*}{\textbf{60.8}} \\
& & & & \color{red}{(-21.6)} && & \color{red}{(-24.8)} && & \color{red}{(-14.6)} &\\
\noalign{\vskip 0.5ex}\cdashline{2-12}\noalign{\vskip 0.5ex}

& \multirow{2}{*}{\begin{tabular}[c]{@{}l@{}}Column Masking\end{tabular}} &
\multirow{2}{*}{177} && 65.2 / 62.3 & \multirow{2}{*}{\textbf{89.6}} & & 68.3 / 65.0 & \multirow{2}{*}{87.4} & & 51.3 / 49.3 & \multirow{2}{*}{\textbf{89.6}} \\
& & & & \color{red}{(-2.9)} && & \color{red}{(-3.3)} && & \color{red}{(-2.0)} &\\
\noalign{\vskip 0.5ex}\cdashline{2-12}\noalign{\vskip 0.5ex}

& \multirow{2}{*}{\begin{tabular}[c]{@{}l@{}}Column Adding\end{tabular}} &
\multirow{2}{*}{2,254} && 62.7 / 60.8 & \multirow{2}{*}{\textbf{92.7}} & & 67.3 / 58.9 & \multirow{2}{*}{81.9} & & 50.2 / 48.5 & \multirow{2}{*}{91.6}\\
& & & & \color{red}{(-1.9)}&& & \color{red}{(-8.4)} && & \color{red}{(-1.7)} &\\

\midrule

\multirow{4}{*}{\begin{tabular}[c]{@{}l@{}}NLQ\end{tabular}} 
& \multirow{2}{*}{\begin{tabular}[c]{@{}l@{}}Word-Level Paraphrase\end{tabular}} &
\multirow{2}{*}{1,198} && 63.6 / 57.7 & \multirow{2}{*}{86.1} & & 68.5 / 63.1 & \multirow{2}{*}{86.9} & & 50.4 / 49.8 & \multirow{2}{*}{\textbf{95.2}} \\
& & & & \color{red}{(-5.9)} && & \color{red}{(-5.4)} && & \color{red}{(-0.6)} &\\
\noalign{\vskip 0.5ex}\cdashline{2-12}\noalign{\vskip 0.5ex}

& \multirow{2}{*}{\begin{tabular}[c]{@{}l@{}}Sentence-Level Paraphrase\end{tabular}} &
\multirow{2}{*}{1,084} && 62.8 / 57.5 & \multirow{2}{*}{85.7}  & & 68.0 / 61.9 & \multirow{2}{*}{86.0} & & 49.8 / 49.5 & \multirow{2}{*}{\textbf{96.3}} \\
& & & & \color{red}{(-5.3)} && & \color{red}{(-6.1)} && & \color{red}{(-0.3)} &\\

\bottomrule
\end{tabular}
}
\caption{Data statistics and robustness evaluation results of state-of-the-art Table QA models on \ours-SQA. \textsc{Acc} represents the \emph{Pre-} and \emph{Post-perturbation Accuracy}; \textsc{R-Acc} represents the \emph{Robustness Accuracy}. Due to the time constraint, we did not construct the \emph{mix} set for \ours-SQA.}
\label{tab:sqa_results}
\end{table*}
\begin{figure*}[!t]
    \centering
    \includegraphics[width = 0.8\linewidth]{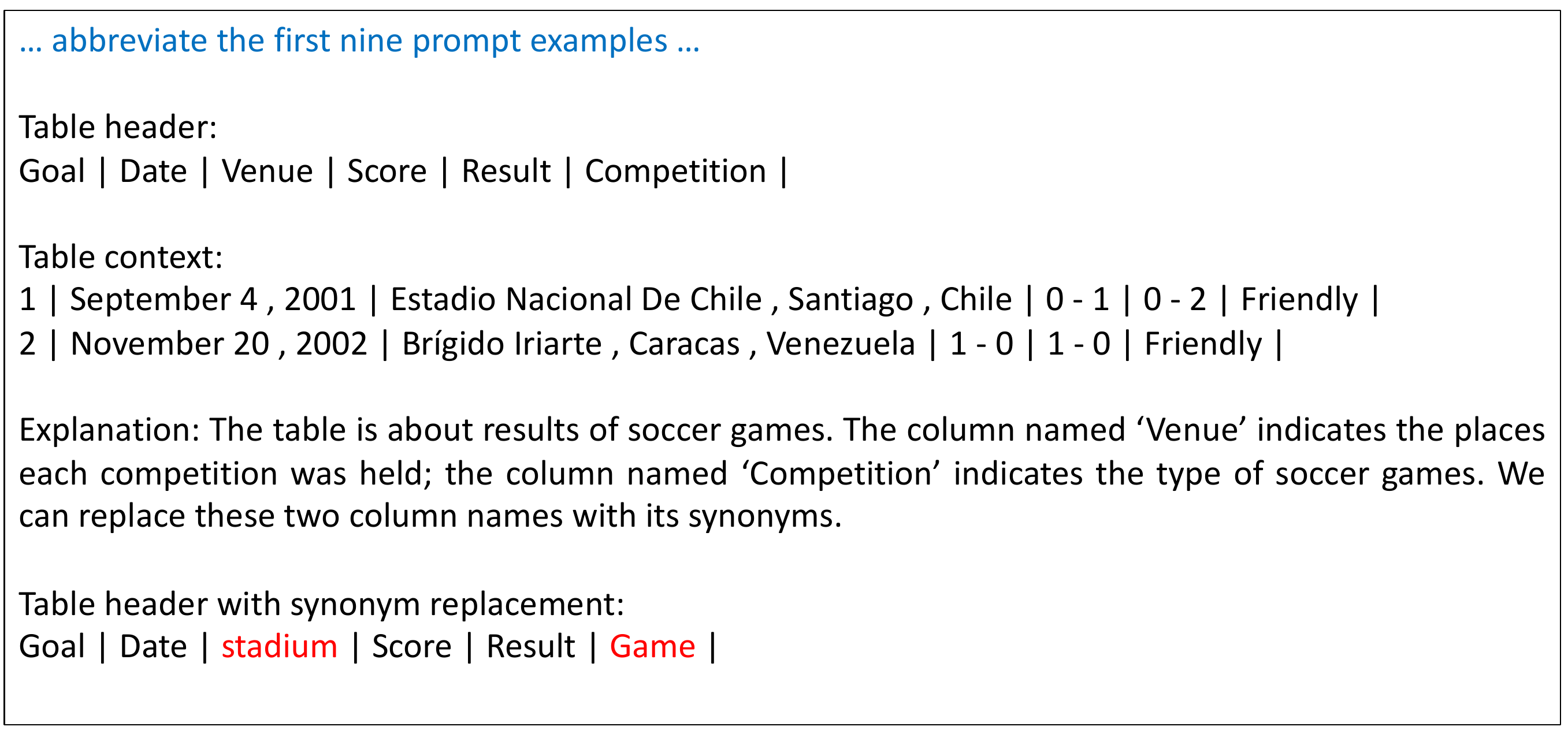}
    \caption{An example of prompt prefix for \emph{header synonym replacement} using GPT-3. The GPT-3 model is prompted to perturb the table header, given the table context (i.e., table header, and first two rows of the table).}
    \label{fig:header_prompt}
\end{figure*}

\begin{figure*}[!t]
    \centering
    \includegraphics[width = 0.8\linewidth]{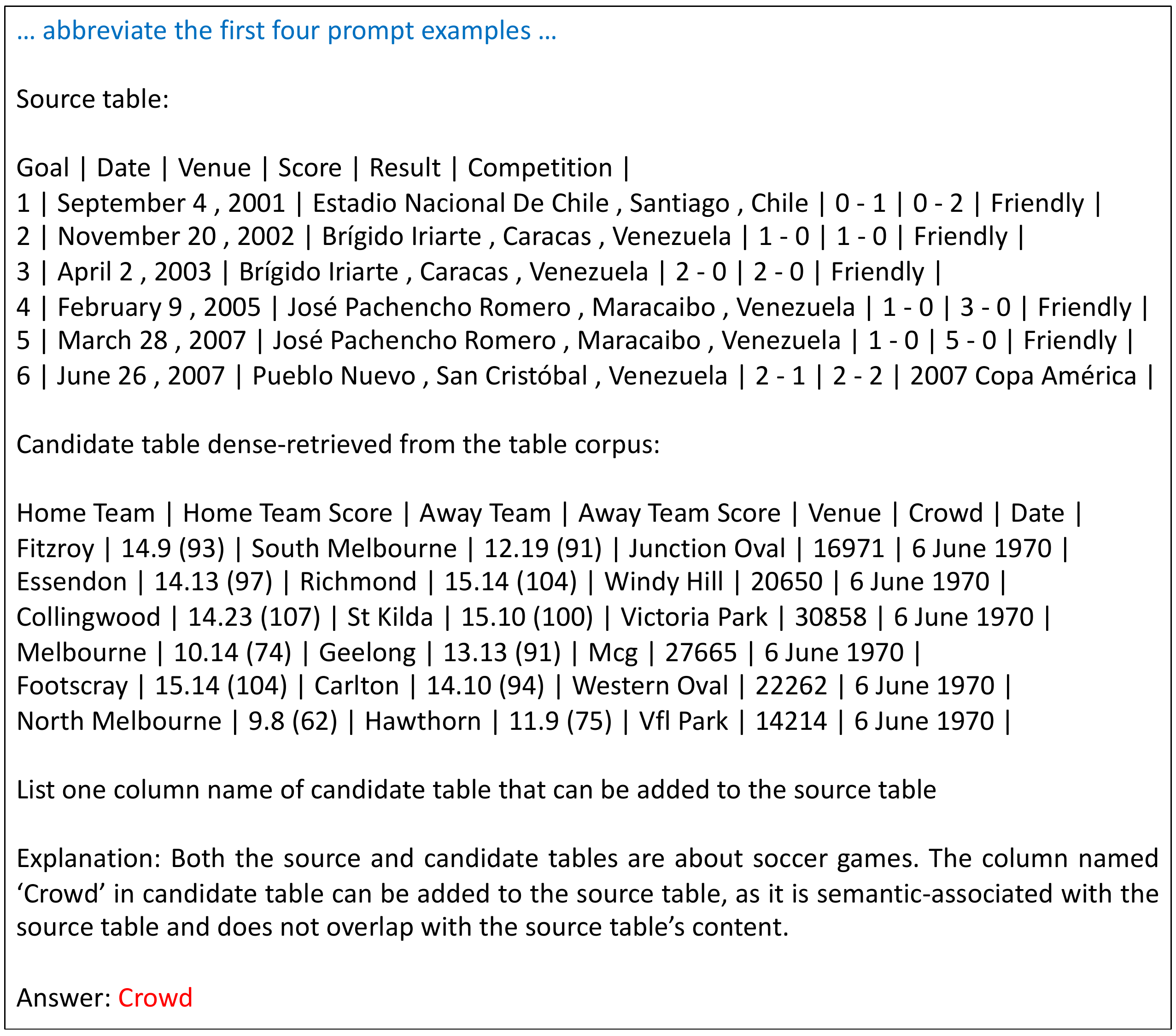}
    \caption{An example of prompt prefix for \emph{column adding} perturbation using CodeX. The candidate table is retrieved by the \tapas-based dense retriever. The CodeX model is prompted to select one column from the candidate table that can be inserted into the source table.}
    \label{fig:table_prompt}
\end{figure*}
\begin{table*}[t!]
\small
\begin{tabular}{p{1cm}p{4.5cm}p{9cm}}
\toprule
Level & Paraphrase Category & Paraphrased Example \\
\midrule
\multirow{16}{*}{Word}
& \texttt{Reasoning-synonym} \newline Paraphrase reasoning operation indicators with its synonyms
& Original: Which was the \red{first} Chinese star map known to have been created?\newline
Paraphrased: Which was the \blue{earliest} Chinese star map known to have been created?
\\\noalign{\vskip 0.5ex}\cdashline{2-3}\noalign{\vskip 0.5ex}

& \texttt{Reasoning-carrier} \newline Rewrite the carrier phrases that are used to infer the reasoning operation
& Original: \red{How many} cities are above 1 million in population\newline
Paraphrased: \blue{What is the quantity of} cities that are above 1 million in population?
\\\noalign{\vskip 0.5ex}\cdashline{2-3}\noalign{\vskip 0.5ex}

& \texttt{Header-synonym} \newline Paraphrase table header indicators with its synonyms
& Original: Who \red{had more points}, Takaji Mori or Junji Kwano?\newline 
Paraphrased: Who \blue{performed better}, Takaji Mori or Junji Kwano? \newline
Explanation: \texttt{points} is the table header name. \\\noalign{\vskip 0.5ex}\cdashline{2-3}\noalign{\vskip 0.5ex}

& \texttt{Header-carrier} \newline Rewrite the carrier phrases used to infer the relevant table columns
& Original: \red{What are the names of players} that scored more than 5 points.\newline 
Paraphrased: \blue{Which athletes} scored more than 5 points? \newline
Explanation: \texttt{Player Name} is the table header name. \\\noalign{\vskip 0.5ex}\cdashline{2-3}\noalign{\vskip 0.5ex}

& \texttt{Cell-Value-synonym} \newline Paraphrase cell value indicators with its synonyms
& Original: How many districts were created in the \red{1900's}?\newline 
Paraphrased: How many districts were created in the \red{twentieth century}? \\\midrule

\multirow{13}{*}{Sentence} & \texttt{Simplification} \newline Simplify the question and make it less redundant 
& Original: How many weeks did the song "Don't Cry for Me Argentina" \red{written by Julie Covington} spend \red{at the first place of} Australia's singles chart?\newline 
Paraphrased: How many weeks was \blue{Julie Covington's} "Don't Cry for Me Argentina" \blue{number one} in Australia's singles chart?? \\\noalign{\vskip 0.5ex}\cdashline{2-3}\noalign{\vskip 0.5ex}

& \texttt{Interrogative Transformation} \newline Convert the question between interrogative and imperative form
& Original: \red{When was} the first game that Kansas State won by double digits? \newline 
Paraphrased: \blue{Please provide me with} the date when Kansas State won the first game by double digits. \\\noalign{\vskip 0.5ex}\cdashline{2-3}\noalign{\vskip 0.5ex}

& \texttt{General} \newline Paraphrase the question in a general way, which might cover multiple paraphrased categories
& Original: \red{What are} the names and \red{stock codes} of companies whose \red{headquarters} are located in the United States? \newline 
Paraphrased: \blue{List} the names and \blue{ticker symbols} of companies \blue{based} in the United States? \\

\bottomrule
\end{tabular}
\caption{Examples of paraphrase categories for \framework NLQ Augmentation. The red words in the original questions highlight the text that are paraphrased. The blue words in the paraphrases represent how the text are replaced.}
\label{tab:nlq_example}
\end{table*}
\begin{figure*}[!t]
    \centering
    \includegraphics[width = 0.8\linewidth]{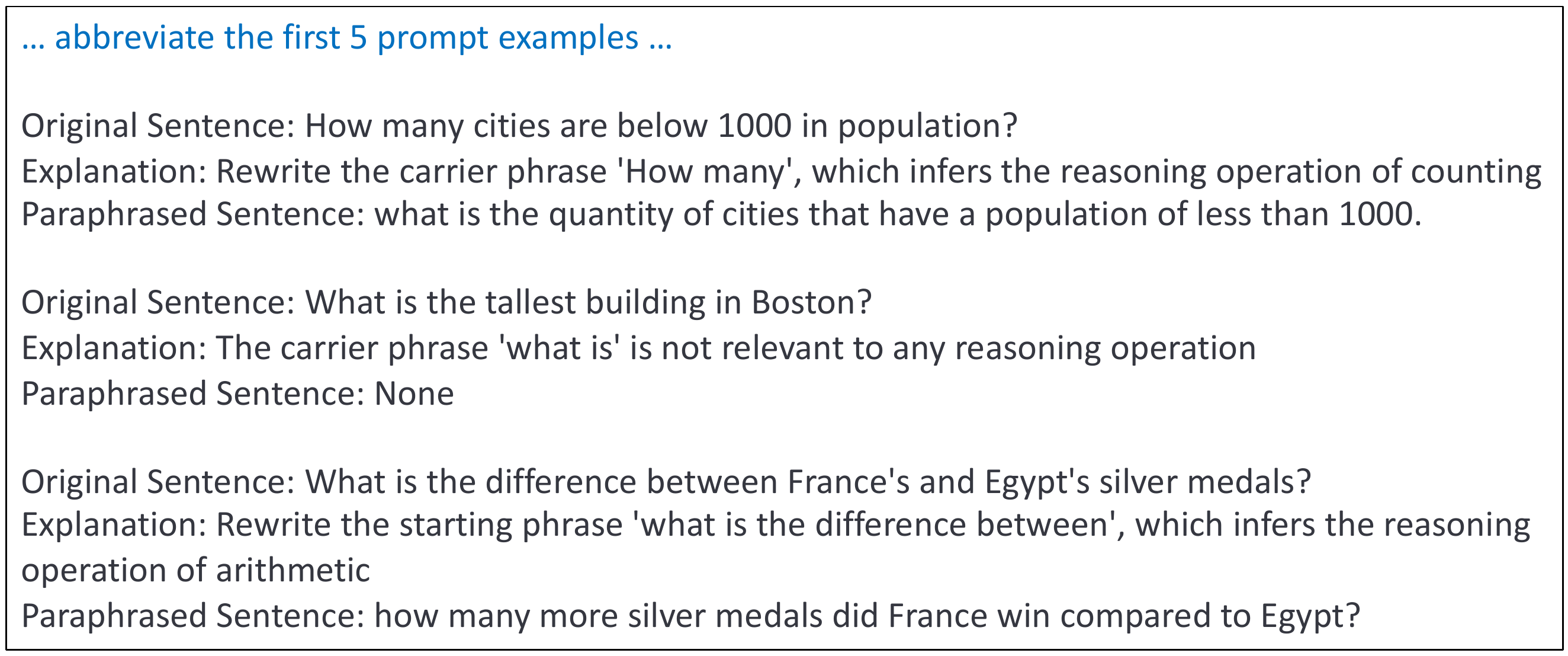}
    \caption{An example of prompt prefix for paraphrasing NLQ with \texttt{Reasoning-synonym} category. For each paraphrase category at the word or sentence level, we designed a demonstration with five to eight examples, where each example includes the original question, the paraphrased question, and corresponding explanations to prompt GPT-3 for generating new paraphrased questions.}
    \label{fig:GPT3qa}
\end{figure*}

\end{document}